\documentclass{article}

   \usepackage[preprint,nonatbib]{neurips_2019}

\usepackage[utf8]{inputenc} 
\usepackage[T1]{fontenc}    
\usepackage{hyperref}       
\usepackage{url}            
\usepackage{booktabs}       
\usepackage{amsfonts}       
\usepackage{nicefrac}       
\usepackage{microtype}      

\usepackage{mathtools}
\usepackage{subcaption}
\usepackage{algorithm}
\usepackage{algorithmicx}
\usepackage[noend]{algpseudocode}
\newcommand{\Tff}{\ensuremath{\Theta_\mathrm{ff}}}
\newcommand{\Tbp}{\ensuremath{\Theta_\mathrm{bp}}}    

\title{SpikeGrad: An ANN-equivalent Computation Model for Implementing Backpropagation with Spikes}

\author{
  Johannes C.~Thiele \\
  CEA, LIST\\
  91191 Gif-sur-Yvette CEDEX, France \\
  \texttt{johannes.thiele@cea.fr} \\
  \And
  Olivier ~Bichler \\
  CEA, LIST\\
  91191 Gif-sur-Yvette CEDEX, France \\
  \texttt{olivier.bichler@cea.fr} \\
  \And
  Antoine ~Dupret \\
  CEA, LIST\\
  91191 Gif-sur-Yvette CEDEX, France \\
  \texttt{antoine.dupret@cea.fr} \\
}

\begin{document}

\maketitle

\begin{abstract}

Event-based neuromorphic systems promise to reduce the energy consumption of deep learning tasks by replacing expensive floating point operations on dense matrices by low power sparse and asynchronous operations on spike events. While these systems can be trained increasingly well using approximations of the backpropagation algorithm, these implementations usually require high precision errors for training and are therefore incompatible with the typical communication infrastructure of neuromorphic circuits. In this work, we analyze how the gradient can be discretized into spike events when training a spiking neural network. To accelerate our simulation, we show that using a special implementation of the integrate-and-fire neuron allows us to describe the accumulated activations and errors of the spiking neural network in terms of an equivalent artificial neural network, allowing us to largely speed up training compared to an explicit simulation of all spike events. This way we are able to demonstrate that even for deep networks, the gradients can be discretized sufficiently well with spikes if the gradient is properly rescaled. This form of spike-based backpropagation enables us to achieve equivalent or better accuracies on the MNIST and CIFAR10 dataset than comparable state-of-the-art spiking neural networks trained with full precision gradients. The algorithm, which we call \textit{SpikeGrad}, is based on accumulation and comparison operations and can naturally exploit sparsity in the gradient computation, which makes it an interesting choice for a spiking neuromorphic systems with on-chip learning capacities.

\end{abstract}

\section{Introduction}

Spiking neural networks (SNNs) are a new generation of artificial neural network models \cite{Maass:1997}, which try to harness potentially useful properties of biological neurons for energy efficient neuromorphic systems. In traditional artificial neural networks (ANNs), processing is based on operations on dense, real valued tensors. In contrast to this, SNNs communicate with asynchronous spike events, which potentially allows them to process efficiently information with high temporal and spatial sparsity if implemented in custom event-based hardware (see for instance \cite{Merolla:2014} and \cite{Qiao:2015}). 

\paragraph{Previous work on optimizing SNNs with backpropagation}
 
The recent years have seen a large number of approaches devoted to optimization of spiking neural networks with the backpropagation algorithm, either by converting ANNs to SNNs \cite{Diehl:2015SpikeConversion}\cite{OConnor:2016}\cite{Esser:2016}\cite{Rueckauer:2017} or by simulating spikes explicitly in the forward pass and optimizing these dynamics with full precision gradients \cite{Lee:2016}\cite{Yin:2017}\cite{Wu:2018SpikingBP}\cite{Wu:2018SpikingBPDeep}\cite{Severa:2019}\cite{Jin:2018}\cite{Belec:2018}\cite{Zenke:2018SuperSpike}. These methods do usually not communicate gradients as spike signals (for a recent and more detailed review of training algorithms for SNNs, see \cite{Pfeiffer:2018} or \cite{Tavanaei:2019}). It would however be desirable to enable on-chip learning in neuromorphic chips using the power of the backpropagation algorithm, while maintaining the advantages of spike-based processing also in the backpropagation phase. Recent work of \cite{Binas:2016} and \cite{Wu:2019} has discussed how forward processing in an SNN could be mapped to an ANN. Our work extends this analysis to the backward propagation pass, to yield a fully spike-based implementation of the backpropagation algorithm.

\paragraph{Previous work on approximating backpropagation with spikes}

In the works of \cite{Neftci:2017} and \cite{Samadi:2017} a spike-based version of the backpropagation algorithm is implemented, using direct feedback to neurons via spike propagation through fixed weights to each layer of the network. While good performance on the MNIST dataset is achieved, they do not demonstrate the capacity of their algorithm on large ANNs and more realistic benchmarks. The exact backpropagation algorithm, which backpropagates through symmetric weights might however be required to reach good performance on large-scale problems \cite{Baldi:2016}\cite{Bartunov:2018}. \cite{Thiele:2019} uses an approximation of the backpropagation algorithm where the error is propagated via spike events to train a network for relational inference. However, no mathematical analysis of the approximate capacities of the algorithm is provided and no scalability to large scale classification problems is demonstrated.

\paragraph{Paper contributions}

Our contributions are twofold: First, we demonstrate how backpropagation can be seamlessly integrated into the spiking neural network framework by using a second accumulation compartment for error propagation, which discretizes the error into spikes. This way we obtain a system that is able to perform learning and inference based on accumulations and comparisons alone. As for the forward pass, this allows us to exploit the dynamic precision and sparsity provided by the discretization of all operations into asynchronous spike events. Secondly, we show that the system obtained in this way can be mapped to an ANN with equivalent accumulated responses in all layers. This allows us to simulate training of large-scale SNNs efficiently on graphic processing units (GPUs), using their equivalent ANN. We demonstrate classification accuracies equivalent or superior to existing implementations of SNNs trained with full precision gradients, and comparable to the precision of standard ANNs. Based on our review of the literature, our work provides for the first time an analysis of how the sparsity of the gradient during backpropagation can be exploited within a large-scale SNN processing structure. This is the first time competitive classification performances are reported on a large-scale spiking network where training and inference are fully implemented with spikes.

\section{The \textit{SpikeGrad} algorithm} \label{sec:acc_based_model}

We begin with the description of \textit{SpikeGrad}, the spike-based backpropagation algorithm. For each training example/mini-batch, integration is performed from $t=0$ to $t=T$ for the forward pass and from $t=T+\Delta t$ to $t=\mathcal{T}$ in the backward pass. Since no explicit time is used in the algorithm, $\Delta t$ represents symbolically the (very short) time between the arrival of an incoming spike and the response of the neuron, which is only used here to describe causality.
 
\paragraph{Integrate-and-fire neuron model}

Our architecture consists of multiple layers (labeled by $l\in [0,L]$) of integrate-and-fire (IF) neurons with integration variable $V_i^l(t)$ and threshold $\Tff$:
\begin{equation} \label{update_potential}
V_i^l(t+ \Delta t) =  V_i^l(t) - \Tff s_i^l(t) + \sum_j w^{l}_{ij}s^{l-1}_j(t),\quad V_i^l(0) = b^l_i.
\end{equation}
The variable $w^{l}_{ij}$ is the weight and $b^l_i$ a bias value. The spike activation function  $s^l_i(t) \in \{-1,0,1\}$ is a function which triggers a signed spike event depending on the internal variables of the neuron. It will be shown later that the specific choice of the activation function is fundamental for the mapping to an equivalent ANN. After a neuron has fired, its integration variable is decremented or incremented by the threshold value $\Tff$, which is represented by the second term on the r.h.s.\ of \eqref{update_potential}.  

As a representation of the neuron activity, we use a trace $x^l_i(t)$ which accumulates spike information over a single example:
\begin{equation} \label{learning_rate_trace}
x_i^{l}(t+\Delta t) = x_i^{l}(t) + \eta s_i^{l}(t).
\end{equation}
By weighting the activity with the learning rate $\eta$ we avoid performing a multiplication when weighting the input with the learning rate for the weight update \eqref{weight_update_final}.

\paragraph{Implementation of implicit ReLU and surrogate activation function derivative}

It is possible to define an implicit activation function based on how the neuron variables affect the spike activation function $s^l_i(t)$. In our implementation, we use the following fully symmetric function to represent linear activation functions (used for instance in pooling layers):
\begin{equation} \label{spike_activation_function_linear}
s^{l,\mathrm{lin}}_i\left(V_i^l(t)\right) \coloneqq \begin{cases}
	\; 1 &  \mathrm{if}\;V^l_i(t)\geq\Tff \\
	\; -1 &  \mathrm{if}\;V^l_i(t)\leq-\Tff\\\
	\; 0 & \mathrm{otherwise} 
	\end{cases}.
\end{equation}
The following function corresponds to the rectified linear unit (ReLU) activation function:
\begin{equation} \label{spike_activation_function_ReLU}
s^{l,\mathrm{ReLU}}_i\left(V_i^l(t),x^l_i(t)\right) \coloneqq \begin{cases}
	\; 1 &  \mathrm{if}\;V^l_i(t)\geq\Tff \\
	\; -1 &  \mathrm{if}\;V^l_i(t)\leq-\Tff\;\mathrm{and}\; x^l_i(t)>0\\
	\; 0 & \mathrm{otherwise} 
	\end{cases}.
\end{equation}

The pseudo-derivative of the activation function is denoted symbolically by $S^{'l}_i$. We use $S_i^{'l,\mathrm{lin}}(T)=1$ for the linear case. For the ReLU, we use a surrogate of the form:
\begin{equation} \label{straight_through_estimator_x}
S_i^{'l,\mathrm{ReLU}}(T) \coloneqq \begin{cases}
	\; 1 &  \mathrm{if}\;  V^l_i(T) > 0 \;\mathrm{or}\;x_i^l(T)>0 \\
	\; 0 & \mathrm{otherwise}
	\end{cases}.
\end{equation}
These choices will be motivated in the following sections. Note that the derivatives depend only on the final states of the neurons at time $T$.

\paragraph{Discretization of gradient into spikes}

For gradient backpropagation, we introduce a second compartment with threshold $\Tbp$ in each neuron, which integrates error signals from higher layers. The process discretizes errors in the same fashion as the forward pass discretizes an input signal into a sequence of signed spike signals:
\begin{equation} \label{update_error}
U_i^l(t+\Delta t) =  U_i^l(t) - \Tbp z_i^l(t) + \sum_k w^{l+1}_{ki}\delta^{l+1}_k(t).
\end{equation}
To this end, we introduce a ternary \textit{error spike activation function} $z_i^{l}(t)\in \{-1,0,1\}$ which is defined in analogy to \eqref{spike_activation_function_linear} using the error integration variable $U_i^l(t)$ and the backpropagation threshold $\Tbp$. The error is then obtained by gating this ternarized variable $z_i^{l}(t)$ with one of the surrogate activation function derivatives of the previous section (linear or ReLU):
\begin{equation}\label{ternarized error}
\delta_i^{l}(t) = z_i^{l}(t) S^{'l}_i(T).
\end{equation}
This ternary spike signal is backpropagated through the weights to the lower layers and also applied in the update rule of the weight increment accumulator $\omega_{ij}^{l}$: 
\begin{equation}\label{weight_update_final}
\omega_{ij}^{l}(t+\Delta t) = \omega_{ij}^{l}(t) -\delta_i^{l}(t)x_j^{l-1}(T),
\end{equation}
which is triggered every time an error spike signal \eqref{ternarized error} is backpropagated. The weight updates are accumulated during error propagation and are applied after propagation is finished to update each weight simultaneously. In this way, the backpropagation of errors and the weight update will, exactly as forward propagation, only involve additions and comparisons of floating point numbers. 

The \textit{SpikeGrad} algorithm can also be expressed in an event-based formulation, described in algorithms \ref{alg1}, \ref{alg2} and \ref{alg3}. This formulation is closer to how the algorithm would be implemented in an actual SNN hardware system.

\paragraph{Loss function and error scale}

We use the cross entropy loss function in the final layer applied to the softmax of the total integrated signal $V^L_i(T)$ (no spikes are triggered in the top layer during inference). This requires more complex operations than accumulations, but is negligible if the number of classes is small. To make sure that sufficient error spikes are triggered in the top layer, and that error spikes arrive even in the lowest layer of the network, we apply a scaling factor $\alpha$ to the error values before transferring them to $U_i^L$. This scaling factor also implicitly sets the precision of the gradient, since a higher number of spikes means that a large range of values can be represented. To counteract the relative increase of the gradient scale, the learning rates have to be rescaled by a factor $\nicefrac{1}{\alpha}$.

\paragraph{Input encoding}

As pointed out in \cite{Rueckauer:2017} and \cite{Wu:2018SpikingBPDeep}, it is crucial to maintain the full precision of the input image to obtain good performances on complex standard benchmarks with SNNs. One possibility is to encode the input in a large number of spikes \cite{Sengupta:2019}. Another possibility, which has been shown to require a much lower number of spikes in the network, is to multiply the input values directly with the weights of the first layer (just like in a standard ANN). The drawback is that the first layer then requires multiplication operations. The additional cost of this procedure may however be negligible if all other layers can profit from spike-based computation. This problematic does not exist for stimuli which are natively encoded in spikes.

\begin{minipage}{0.49\textwidth}
\begin{algorithm}[H]
\caption{Forward}\label{alg1}
\begin{algorithmic}
\Function{Propagate}{$[l,i,j],s$} 
\State $V^l_i \gets V^l_i + s\cdot w^l_{ij}$
\State $s^l_i \gets s^l_i(V^l_i,x^l_i)$ \Comment{spike activation function}
\If{$s^l_i \neq 0$} 
\State $V^l_i \gets V^l_i - s^l_i\cdot \Tff$
\State $x^l_i \gets x^l_i + \eta s^l_i$
\For{$k$ in $l+1$ connected to $i$}
\State \Call{Propagate}{$[l+1,k,i],s^l_i$}
\EndFor
\EndIf
\EndFunction
\end{algorithmic}
\end{algorithm}
\end{minipage}
\hfill
\begin{minipage}{0.49\textwidth}
\begin{algorithm}[H]
\caption{Backward}\label{alg2}
\begin{algorithmic}
\Function{Backpropagate}{$[l,i,k],\delta$} 
\State $U^l_i \gets U^l_i + \delta\cdot w^{l+1}_{ki}$
\State $z^l_i \gets z^l_i(U^l_i)$ \Comment{error activation function}
\State $\delta^l_i \gets z^l_i \cdot S^{'l}_i$ 
\If{$z^l_i \neq 0$} 
\State $U^l_i \gets U^l_i - z^l_i\cdot \Tbp$
\For{$j$ in layer $l-1$ connected to $i$}
\State \Call{Backpropagate}{$[l-1,j,i],\delta^l_i$}
\State $\omega^l_{ij} \gets \omega^l_{ij} - \delta^l_i\cdot x^{l-1}_j$
\EndFor
\EndIf
\EndFunction
\end{algorithmic}
\end{algorithm}
\end{minipage}

\begin{algorithm}
\caption{Training of single example/batch}\label{alg3}
\begin{algorithmic}
\State \textbf{init:} $\mathbf{V}\gets\mathbf{b}$, $\mathbf{U}\gets 0$, $\mathbf{x}\gets 0$, $\mathbf{\omega}\gets 0$
\Comment{variables in \textbf{bold} describe all neurons in network/layer}
\While{input spikes $s^{in}_i$} 
\For{$k$ in $l=0$ receiving $s^{in}_i$}
\Comment{spikes corresponding to training input}
\State \Call{Propagate}{$[0,k,i],s^{in}_i$}
\EndFor
\EndWhile
\State $\mathbf{S'} \gets \mathbf{S'}(\mathbf{V},\mathbf{x})$ \Comment{calculate surrogate derivatives}
\State $\mathbf{U^L} \gets \alpha \cdot \nicefrac{\partial\mathcal{L}}{\partial\,\mathrm{softmax}(\mathbf{V^L})}$ 
\Comment{calculate classfication error}
\While{$|U_i^L| \geq \Tbp$} 
\Comment{backpropagate error spikes}
\State \Call{Backpropagate}{$[L,i,-],0$}
\Comment{last layer receives no error from higher neurons}
\EndWhile
\State $\mathbf{w} \gets \mathbf{w} + \mathbf{\omega}$
\end{algorithmic}
\end{algorithm}

\section{Formulation of the equivalent ANN}

The simulation of the temporal dynamics of spikes requires a large number of time steps or events if activations are large. It would therefore be extremely beneficial if we were able to map the SNN to an equivalent ANN that can be trained much faster on standard hardware. In this section, we demonstrate that it is possible to find such an ANN using the forward and backward propagation dynamics described in the previous section.  

\paragraph{Spike discretization error}

We start our analysis with equation \eqref{update_potential}. We reorder the terms and sum over the increments $\Delta V_i^l(t) = V_i^l(t+\Delta t) -V_i^l(t)$ every time the integration variable is changed either by a spike that arrives at time $t_j^s \in [0, T]$ via connection $j$, or by a spike that is triggered at time $t_i^s\in [0, T]$. With the initial conditions $V_i^l(0)=b^l_i$, $s_i^l(0)=0$, we obtain the final value $V_i^l(T)$:
\begin{equation} 
V_i^l(T) = \sum_{t_j^s,t_i^s}\Delta V_i^l = - \Tff \sum_{t_i^s} s_i^l(t_i^s) + \sum_j w^{l}_{ij}\sum_{t_j^s} s^{l-1}_j(t_j^s) + b^l_i
\end{equation}
By defining the total transmitted output of a neuron as $S_i^{l} \coloneqq \sum_{t_i^s} s_i^l(t_i^s)$ we obtain:
\begin{equation} \label{total_activation}
\frac{1}{\Tff}V_i^l(T) = \mathbb{S}_i^{l} - S_i^{l},\quad \mathbb{S}_i^{l} \coloneqq \frac{1}{\Tff}\left(\sum_j w^{l}_{ij}S_j^{l-1} + b^l_i\right)
\end{equation}

The same reasoning can be applied to backpropagation of the gradient. We define the summed responses over error spikes times $\tau_j^s \in [T+\Delta t,\mathcal{T}]$ as $Z_i^{l} \coloneqq \sum_{\tau_i^s} z_i^l(\tau_i^s)$ to obtain:
\begin{equation} \label{total_error}
\frac{1}{\Tbp}U_i^l(\mathcal{T}) = \mathbb{Z}_i^{l} - Z_i^{l},\quad \mathbb{Z}_i^{l} \coloneqq \frac{1}{\Tbp}\left(\sum_k w^{l+1}_{ki}E_k^{l+1}\right)
\end{equation}
\begin{equation}
E_k^{l+1} = \sum_{\tau^s_k} \delta_k^{l+1}(\tau^s_k) = \sum_{\tau^s_k} S_k^{'l+1}(T) z_k^{l+1}(\tau^s_k) =  S_k^{'l+1}(T) Z_k^{l+1}.
\end{equation}
In both equation \eqref{total_activation} and \eqref{total_error}, the terms $\mathbb{S}_i^{l}$ and $\mathbb{Z}_i^{l}$ are equivalent to the output of an ANN with signed integer inputs $S^{l-1}_j$ and $E^{l+1}_k$. The scaling factors $\nicefrac{1}{\Tff}$ and $\nicefrac{1}{\Tbp}$ can be interpreted as a linear activation function in the case of the forward pass, and a gradient rescaling in the case of the backward pass. If gradients shall not be explicitly rescaled, backpropagation requires $\Tbp=\Tff$. The values of the residual integrations $\nicefrac{1}{\Tff}V_i^l(T)$ and $\nicefrac{1}{\Tbp}U_i^l(\mathcal{T})$ therefore represent the \textit{spike discretization error} $\mathrm{SDE}_\mathrm{ff} \coloneqq \mathbb{S}_i^{l} - S_i^{l}$ or $\mathrm{SDE}_\mathrm{ff} \coloneqq \mathbb{Z}_i^{l} - Z_i^{l}$ between the ANN outputs $\mathbb{S}_i^{l}$ and $\mathbb{Z}_i^{l}$ and the accumulated SNN outputs $S_i^{l}$ and $Z_i^{l}$. Since we know that $V_i^l(T)\in(-\Tff,\Tff)$ and $U_i^l(\mathcal{T})\in(-\Tbp,\Tbp)$, this gives bounds of $|\mathrm{SDE}_\mathrm{ff}| < 1$ and $|\mathrm{SDE}_\mathrm{bp}| < 1$. 

So far we can only represent linear functions. We now consider an implementation where the ANN applies a ReLU activation function instead. The SDE in this case is: 
\begin{equation} \label{total_activation_integer_ReLU}
\mathrm{SDE}^\mathrm{ReLU}_\mathrm{ff} \coloneqq \mathrm{ReLU}\left(\mathbb{S}^l_i\right) - S_i^{l}.
\end{equation}
We can calculate the error by considering that \eqref{spike_activation_function_ReLU} forces the neuron in one of two regimes (note that $x_i^{l} > 0 \Leftrightarrow S_i^{l} > 0$): In one case, $S_i^{l} = 0,\,V_i^{l}(T) < \Tff$ (this includes $V_i^l(T) \leq -\Tff$). This implies $\mathbb{S}^l_i = \nicefrac{1}{\Tff}V_i^l(T)$ and therefore $|\mathrm{SDE}^\mathrm{ReLU}_\mathrm{ff}| < 1$ (or even $|\mathrm{SDE}^\mathrm{ReLU}_\mathrm{ff}| = 0$ if $V_i^{l}(T) \leq 0$). 
In the other case, $S_i^{l} > 0, \,V_i^{l}(t) \in (-\Tff,\Tff)$, where \eqref{spike_activation_function_ReLU} is equivalent to \eqref{spike_activation_function_linear}. 

This equivalence motivates the choice of \eqref{straight_through_estimator_x} as a surrogate derivative for the SNN: the condition $(V^l_i(T) > 0 \;\mathrm{or}\;x_i^l(T)>0)$ can be seen to be equivalent to $\mathbb{S}^l_i(T) >  0$, which defines the derivative of a ReLU. Finally, for the total weight increment $\Delta w_{ij}^l$, it can be seen from \eqref{learning_rate_trace} and \eqref{weight_update_final} that:
\begin{equation} 
x_i^l(T) = \sum_{t_i^s}\Delta x_i^{l}(t_i^s) = \eta S_i^{l},\quad\Rightarrow \quad\Delta w_{ij}^l(\mathcal{T}) = \sum_{\tau^s_i} \Delta \omega_{ij}^{l}(\tau^s_i) = -\eta S_j^{l-1}E_i^l,
\end{equation}
which is exactly the weight update formula of an ANN defined on the accumulated variables. We have therefore demonstrated that the SNN can be represented by an ANN by replacing recursively all $S$ and $Z$ by $\mathbb{S}$ and $\mathbb{Z}$ and applying the corresponding activation function directly on these variables. The error that will be caused by this substitution compared to using the accumulated variables $S$ and $Z$ of an SNN is described by the SDE. This ANN can now be used for training of the SNN on GPUs. The \textit{SpikeGrad} algorithm formulated on the variables $s$, $z$, $\delta$ and $x$ represents the algorithm that would be implemented on a event-based \textit{spiking} neural network hardware platform. We will now demonstrate how the SDE can be further reduced to obtain an ANN and SNN that are exactly equivalent.

\paragraph{Response equivalence}

For a large number of spikes, the SDE may be negligible compared to the activation of the ANN. However, in a framework whose objective it is to minimize the number of spikes emitted by each neuron, this error can have a potentially large impact.

One option to reduce the error between the ANN and the SNN output is to constrain the ANN during training to integer values. One possibility is to round the ANN outputs:
\begin{equation} \label{total_activation_integer}
\mathbb{S}_i^{l,\mathrm{round}} \coloneqq \mathrm{round}[\mathbb{S}_i^{l}] = \mathrm{round}\left[{\frac{1}{\Tff}\left(\sum_j w^{l}_{ij}S_j^{l-1} + b^l_i\right)}\right],
\end{equation}
The $\mathrm{round}$ function here rounds to the next integer value, with boundary cases rounded \textit{away} from zero. This behavior can be implemented in the SNN by a modified spike activation function which is applied after the full stimulus has been propagated. To obtain the exact response as the ANN, we have to take into account the current value of $S_i^l$ and modify the threshold values:
\begin{equation} \label{spike_activation_residual}
s^{l,\mathrm{res}}_i\left(V_i^l(T), S_i^l\right)  \coloneqq \begin{cases}
	\; 1 &  \mathrm{if}\;V^l_i(T) >\nicefrac{\Tff}{2}\;\mathrm{or}\;(S_i^l \geq 0,\;V^l_i(T) = \nicefrac{\Tff}{2})\\
	\; -1 &  \mathrm{if}\;V^l_i(T) <-\nicefrac{\Tff}{2}\;\mathrm{or}\;(S_i^l \leq 0,\;V^l_i(T) = -\nicefrac{\Tff}{2})\\
	\; 0 & \mathrm{otherwise} 
	\end{cases}.
\end{equation}
Because this spike activation function is applied only to the residual values, we call it the \textit{residual spike activation function}. The function is applied to a layer after all spikes have been propagated with the standard spike activation function \eqref{spike_activation_function_linear} or \eqref{spike_activation_function_ReLU}. We start with the lowest layer and propagate all residual spikes to the higher layers, which use the standard activation function. We then proceed with setting the next layer to residual mode and propagate the residual spikes. This is continued until we arrive at the last layer of the network.

By considering all possible rounding scenarios, it can be seen that \eqref{spike_activation_residual} indeed implies:
\begin{equation} 
S_i^l + s^{l,\mathrm{res}}_i\left(V_i^l(T), S_i^l\right) = \mathrm{round}[S_i^l + \nicefrac{1}{\Tff} V_i^l(T)] = \mathrm{round}[\mathbb{S}_i^{l}].
\end{equation}
The same principle can be applied to obtain integer-rounded error propagation:
\begin{equation} \label{total_error_integer}
\mathbb{Z}_i^{l,\mathrm{round}}  \coloneqq \mathrm{round}\left[\mathbb{Z}_i^{l}\right] = \mathrm{round}\left[\frac{1}{\Tbp}\left(\sum_k w^{l+1}_{ki}E_k^{l+1} \right)\right].
\end{equation}
We have to apply the following modified spike activation function in the SNN after the full error has been propagated by the standard error spike activation function:
\begin{equation} \label{error_activation_residual}
z^{l,\mathrm{res}}_i\left(U_i^l(\mathcal{T},Z_i^l\right)  \coloneqq \begin{cases}
	\; 1 &  \mathrm{if}\;U_i^l(\mathcal{T})) >\nicefrac{\Tbp}{2}\;\mathrm{or}\;(Z_i^l \geq 0,\;U_i^l(\mathcal{T}) = \nicefrac{\Tbp}{2})\\
	\; -1 & \mathrm{if}\;U_i^l(\mathcal{T}) <-\nicefrac{\Tbp}{2}\;\mathrm{or}\;(Z_i^l \leq 0,\;U_i^l(\mathcal{T}) = -\nicefrac{\Tbp}{2})\\
	\; 0 & \mathrm{otherwise}
	\end{cases},
\end{equation}
which implies:
\begin{equation} 
Z_i^l + z^{l,\mathrm{res}}_i\left(U_i^l(\mathcal{T}), Z_i^l\right) = \mathrm{round}[Z_i^l + \nicefrac{1}{\Tbp}U_i^l(\mathcal{T})] = \mathrm{round}[\mathbb{Z}_i^{l}].
\end{equation}
We have therefore shown that the SNN will after each propagation phase have exactly the same accumulated responses as the corresponding ANN. The same principle can be applied to obtain other forms of rounding (e.g.\ floor and ceil), if \eqref{spike_activation_residual} and \eqref{error_activation_residual} are modified accordingly. 

\paragraph{Computational complexity estimation}

Note that we have only demonstrated the equivalence of the accumulated neurons responses. However, for each of the response values, there is a large number of possible combinations of $1$ and $-1$ values that lead to the same response. The computational complexity of the event-based algorithm depends therefore on the total number $n$ of these events. The best possible case is when the accumulated response value $S^l_i$ is represented by exactly $|S^l_i|$ spikes. In the worst case, a large number of additional redundant spikes is emitted which sum up to $0$. The maximal number of spikes in each layer is bounded by the largest possible integration value that can be obtained. This depends on the maximal weight value $w^l_\mathrm{max}$, the number of connections $N^l_\mathrm{in}$ and the number of spike events $n^{l-1}$ each connection receives, which is given by the maximal value of the previous layer (or the input in the first layer): 
\begin{equation}
n_\mathrm{min}^l = |S^l_i|, \quad n^l_\mathrm{max} = \left\lfloor\frac{1}{\Tff} N^l_\mathrm{in} w^l_\mathrm{max} n^{l-1}_\mathrm{max}\right\rfloor.
\end{equation}
The same reasoning applies to backpropagation. Our experiments show that for input encodings where the input is provided in a continuous fashion, and weight values which are much smaller than the threshold value, the deviation from the best case scenario is rather small. This is because in this case the sub-threshold integration allows to average out the fluctuations in the signal. This way the firing rate stays rather close to its long term average and few redundant spikes are emitted. For the total number of spikes $n$ in the full network on the CIFAR10 test set, we obtain empirically $\nicefrac{n - n_\mathrm{min}}{n_\mathrm{min}} < 0.035$.

\section{Experiments}

For all experiments, the means, errors and maximal values are calculated over 20 simulation runs.

\paragraph{Classification performance}

Tables \ref{Compare_performance_MNIST} and \ref{Compare_performance_cifar10}  compare the state-of-the-art results for SNNs on the MNIST and CIFAR10 datasets. It can be seen that in both cases our results are competitive with respect to the state-of-the-art results of other SNNs trained with high precision gradients. Compared to results using the same topology, our algorithm performs at least equivalently. 

\begin{figure}
	\centering
	\begin{subfigure}[b]{0.49\textwidth}
		\centering
		\includegraphics[width=1\textwidth]{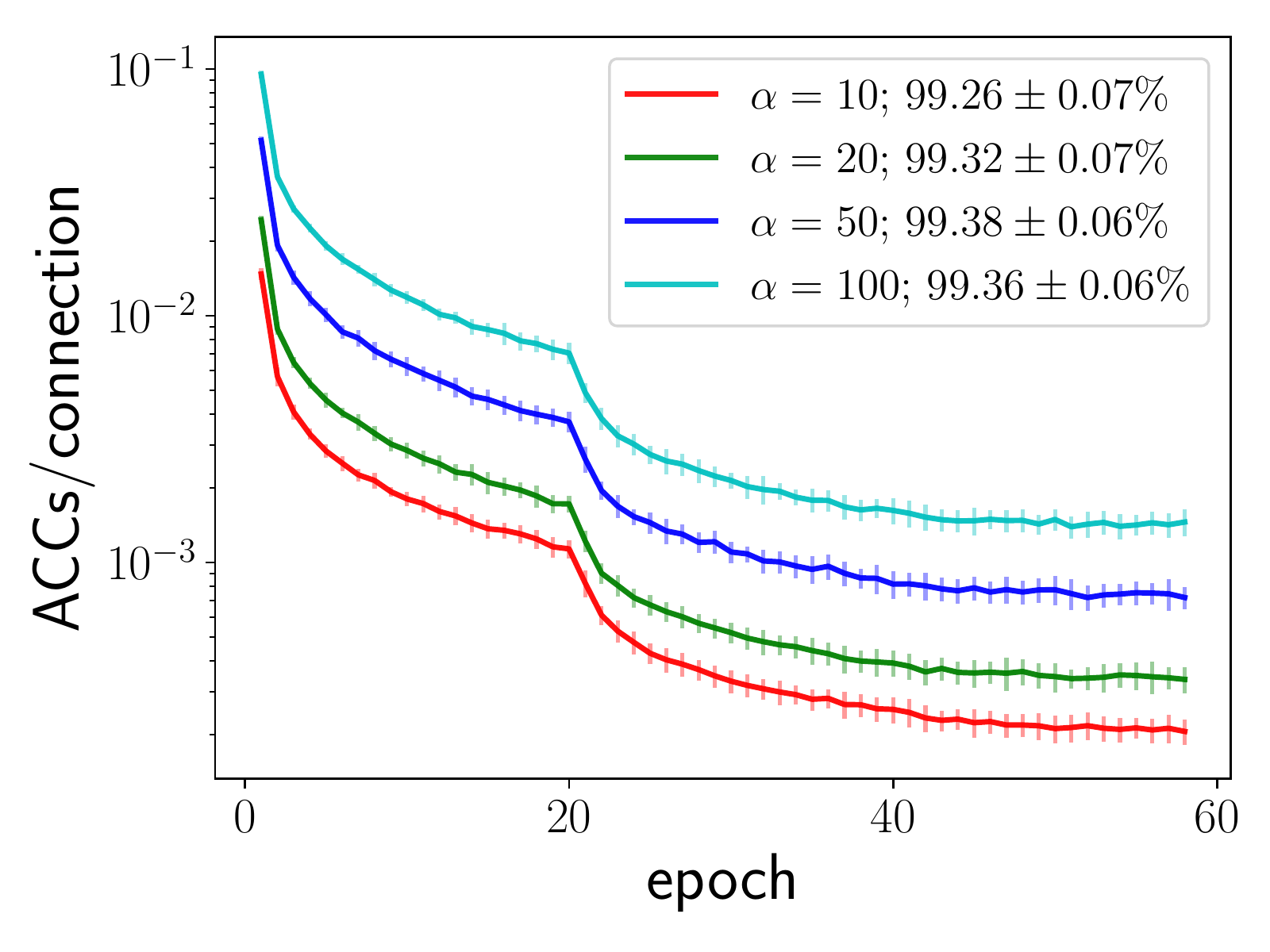}
              \caption{MNIST}
	\label{fig:bp-metrics-scaleVar-MNIST}
	\end{subfigure}
	 \hfill
	\begin{subfigure}[b]{0.49\textwidth}
		\centering
		\includegraphics[width=1\textwidth]{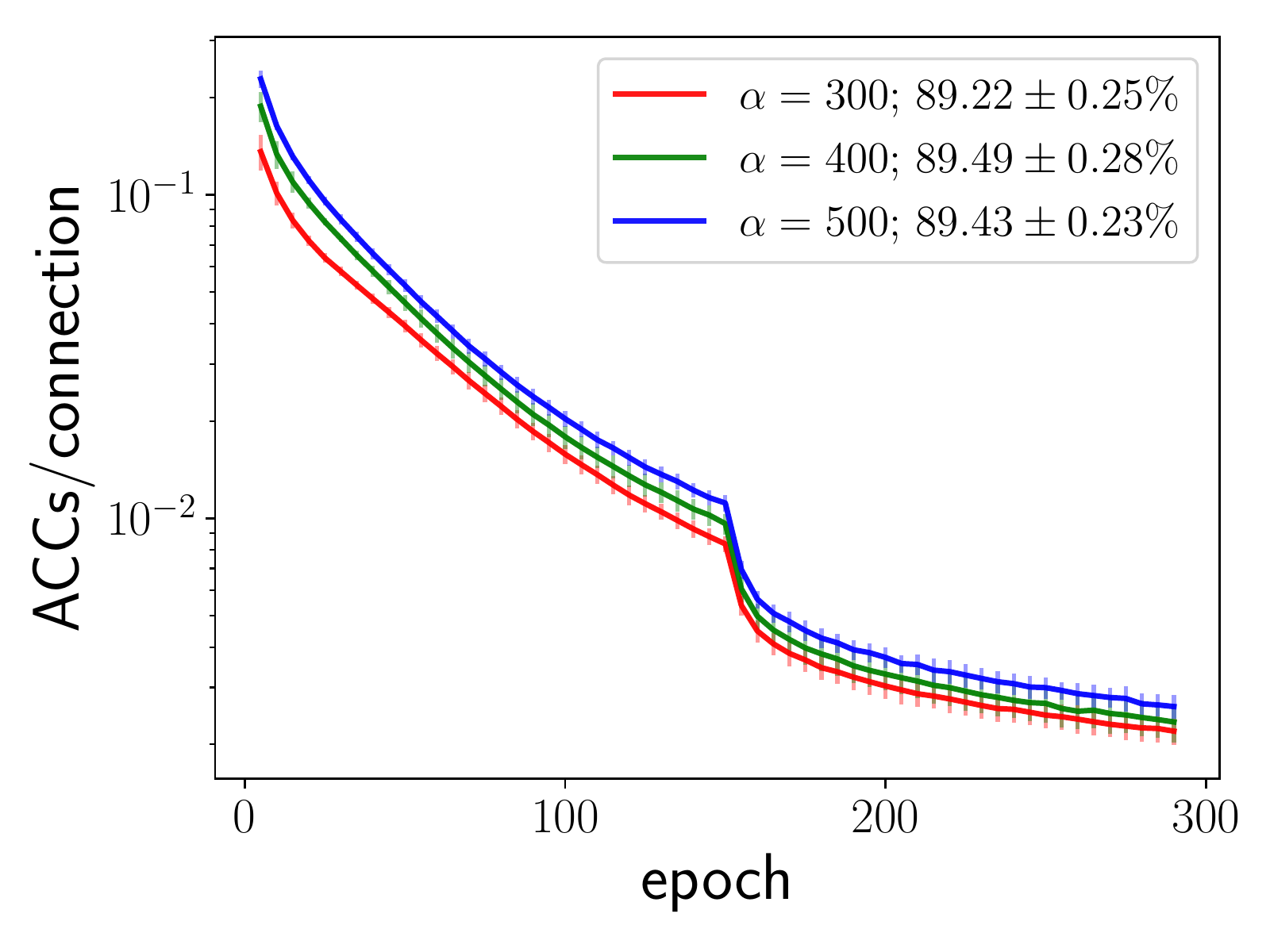}
              \caption{CIFAR10}
	\label{fig:bp-metrics-scaleVar-CIFAR10}
	\end{subfigure}
	\caption{Number of relative synaptic operations during backpropagation for different error scaling factors $\alpha$ as a function of the epoch. Numbers are based on activation values of the equivalent ANN. Test performance with error is given for each $\alpha$.}
    \label{fig:bp-metrics-scaleVar}
\end{figure}

The final classification performance of the network as a function of the error scaling term $\alpha$ in the final layer can be seen in figure \ref{fig:bp-metrics-scaleVar}. Previous work on low bitwidth gradients \cite{Zhou:2018DoReFa} found that gradients usually require a higher precision than both weights and activations. Our results also seem to indicate that a certain minimum number of error spikes is necessary to achieve convergence. This strongly depends on the depth of the network and if enough spikes are triggered to provide sufficient gradient signal in the bottom layers. For the CIFAR10 network, convergence becomes unstable for approximately $\alpha < 300$. If the number of operations is large enough for convergence, the required precision for the gradient does not seem to be extremely large. On the MNIST task, the difference in test performance between a gradient rescaled by a factor of 50 and a gradient rescaled by a factor of 100 becomes insignificant. In the CIFAR10 task, this is true for a rescaling by 400 or 500. Also the results obtained with the float precision gradients in tables \ref{Compare_performance_MNIST} and \ref{Compare_performance_cifar10} demonstrate the same performance, given the range of the error.  

\begin{table}
\caption{Comparison of different state-of-the-art spiking CNN architectures on MNIST. * indicates that the same topology (28x28-15C5-P2-40C5-P2-300-10) was used. }
\label{Compare_performance_MNIST}
\centering
\begin{tabular}{lll}
\toprule
Architecture & Method & Rec. Rate (\textbf{max}[mean$\pm$std]) \\ 
\midrule
Wu et al. \cite{Wu:2018SpikingBP}* & Direct training float gradient & $99.42\%$ \\
Rueckauer et al. \cite{Rueckauer:2017} &CNN converted to SNN & $99.44\%$ \\
Jin et al. \cite{Jin:2018}* & Direct Macro/Micro BP & $99.49\%$ \\
\textbf{This work}* & Direct float gradient & $\mathbf{99.48}[99.36\pm 0.06]\%$ \\
\textbf{This work}* & Direct spike gradient & $\mathbf{99.52}[99.38\pm 0.06]\%$ \\
\bottomrule 
\end{tabular}
\end{table}

\begin{table}
\caption{Comparison of different state-of-the-art spiking CNN architectures on CIFAR10.  * indicates that the same topology (32x32-128C3-256C3-P2-512C3-P2-1024C3-512C3-1024-512-10) was used. }
\label{Compare_performance_cifar10}
\centering
\begin{tabular}{lll}
\toprule
Architecture & Method & Rec. Rate (\textbf{max}[mean$\pm$std])  \\ 
\midrule
Rueckauer et al. \cite{Rueckauer:2017} & CNN converted SNN (with BatchNorm)& $90.85\%$ \\
Sengupta et al. \cite{Sengupta:2019} & VGG-16 converted to SNN& $91.55\%$ \\
Wu et al. \cite{Wu:2018SpikingBPDeep}* & Float gradient (no NeuNorm) & $89.32\%$ \\
\textbf{This work}* & Direct float gradient & $\mathbf{89.72}[89.38\pm 0.25]\%$ \\
\textbf{This work}* & Direct spike gradient & $\mathbf{89.99}[89.49\pm 0.28]\%$ \\
\bottomrule 
\end{tabular}
\end{table}

\paragraph{Sparsity in backpropagated gradient}

To evaluate the potential efficiency of the spike coding scheme relative to an ANN, we use the metric of relative synaptic operations. A synaptic operation corresponds to a multiply-accumulate (MAC) in the case of an ANN, and a simple accumulation (ACC) in the case of an SNN. This metric allows us to compare networks based on their fundamental operation. The advantage of this metric is the fact that it does not depend on the exact implementation of the operations (for instance the number of bits used to represent each number). Since an ACC is however generally cheaper and easier to implement than a MAC, we can be sure that an SNN is more efficient in terms of its operations than the corresponding ANN if the number of ACCs is smaller than the number of MACs. 

\begin{figure}
	\centering
	\begin{subfigure}[b]{0.49\textwidth}
		\centering
		\includegraphics[width=1\textwidth]{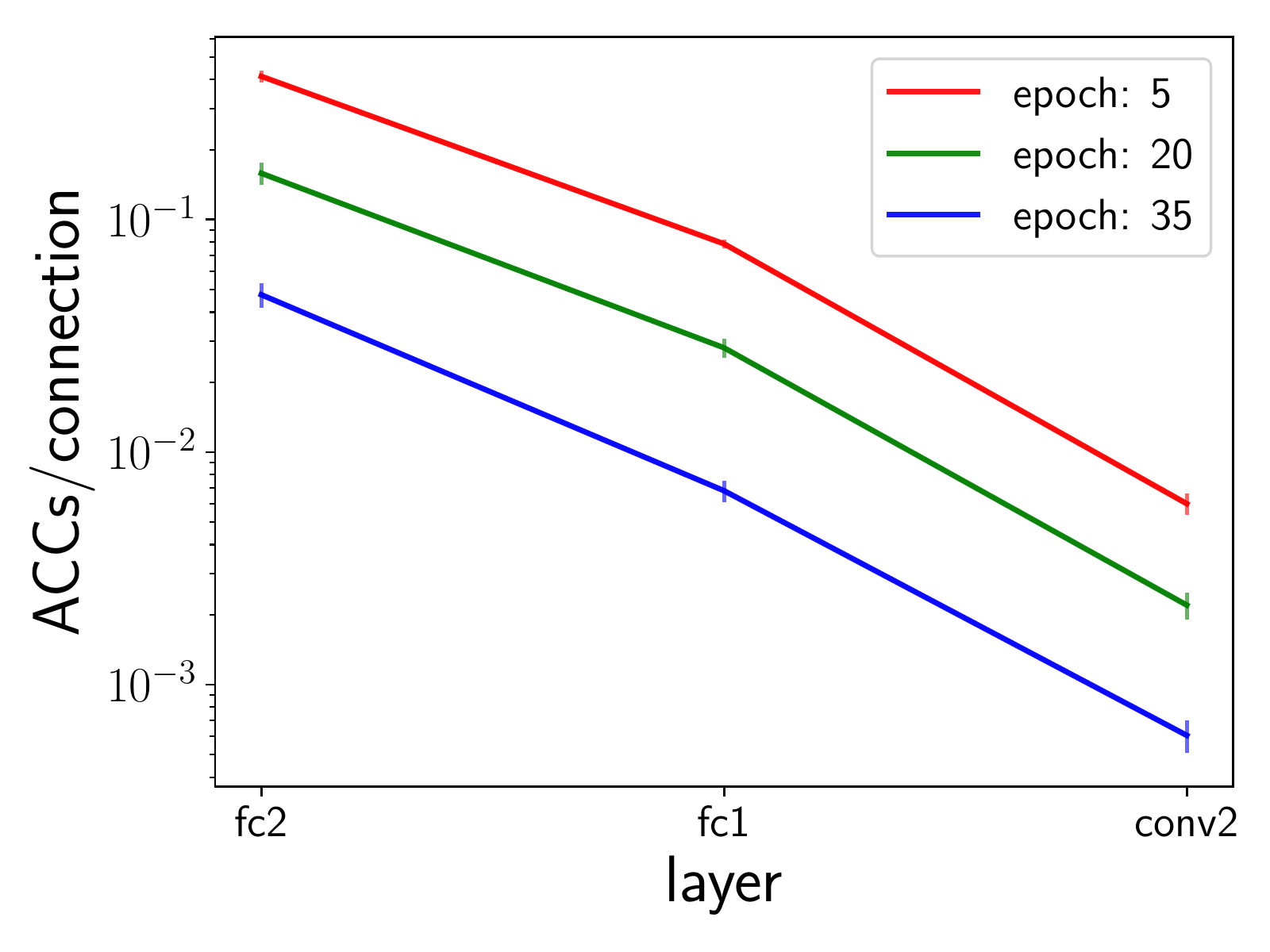}
              \caption{MNIST}
	\label{fig:operations_layer_bp_forward_MNIST}
	\end{subfigure}
	 \hfill
	\begin{subfigure}[b]{0.49\textwidth}
		\centering
		\includegraphics[width=1\textwidth]{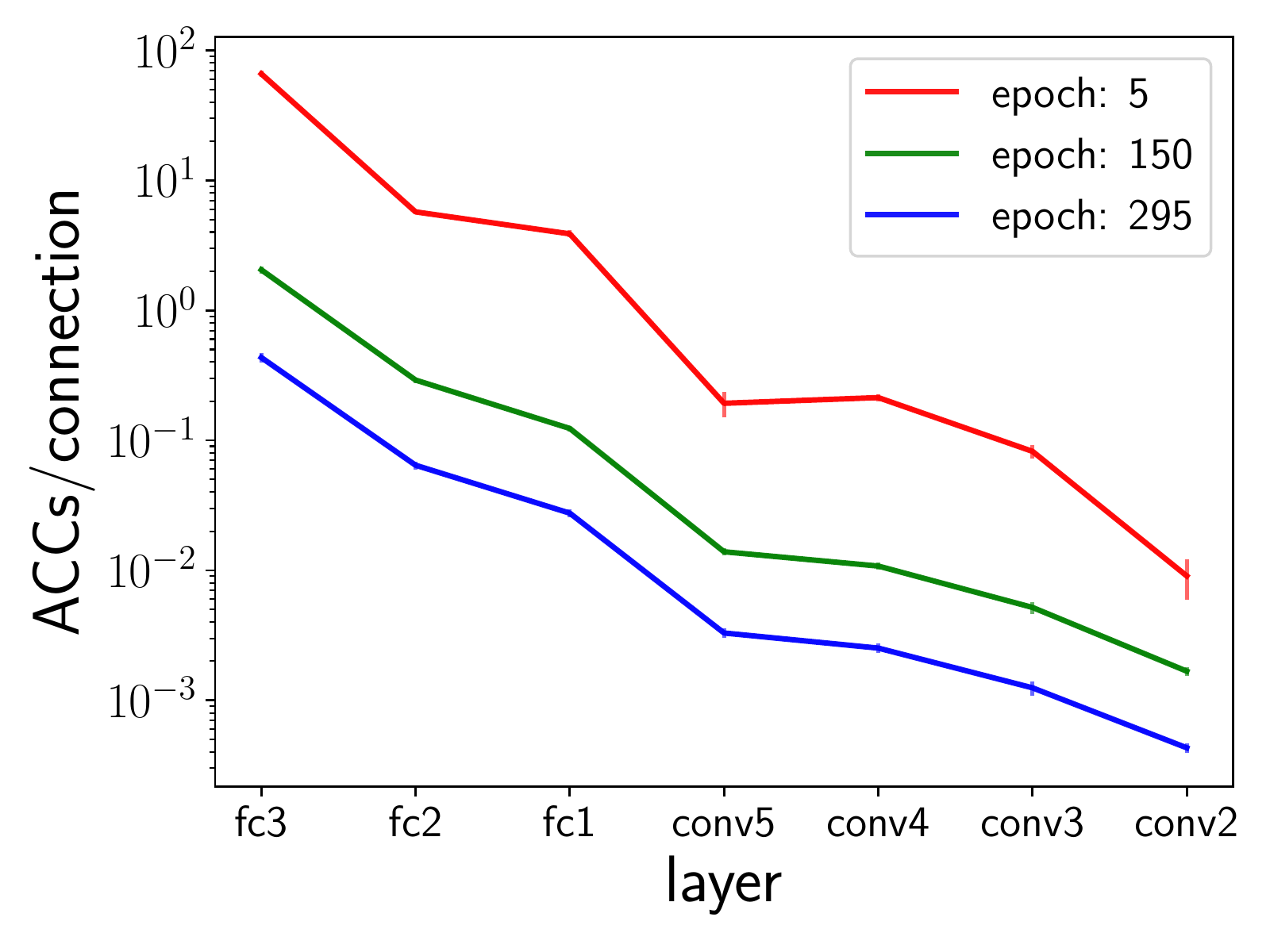}
              \caption{CIFAR10}
	\label{fig:operations_layer_bp_forward_CIFAR10}
	\end{subfigure}
	\caption{Number of relative synaptic operations during backpropagation in each layer (connections in direction of backpropagation) for different epochs. For MNIST $\alpha=100$, for CIFAR10 $\alpha=500$.}
    \label{fig:operations_layer_bp_forward}
\end{figure}

In figure \ref{fig:bp-metrics-scaleVar} it can be seen that the number of operations (i.e.\ the number of spikes) decreases with increasing inference precision of the network. This is a result of the decrease of error in the classification layer, which leads to the emission of a smaller number of error spikes. Numbers were obtained with the integer activation of the equivalent ANN to keep simulation times tractable. As explained previously, the actual number of events and synaptic operations in an SNN may therefore slightly deviate from these numbers. Figure \ref{fig:operations_layer_bp_forward} demonstrates how the number of operations during the backpropagation phase is distributed in the layers of the network (float precision input layer and average pooling layers were omitted). While propagating deeper into the network, the relative number of operations decreases and the error becomes increasingly sparse. This tendency is consistent during the whole training process for different epochs. 


\section{Discussion and conclusion}

Using spike-based propagation of the error gradient, we demonstrated that the paradigm of event-based information propagation can be translated to the backpropagation algorithm. We have not only shown that competitive inference performance can be achieved, but also that gradient propagation seems particularly suitable to leverage spike-based processing by exploiting high signal sparsity. For both forward and backward propagation, \textit{SpikeGrad} requires a similar communication infrastructure between neurons, which simplifies a possible spiking hardware implementation. One restriction of our algorithm is the need for negative spikes, which could be problematic depending on the particular hardware implementation.

In particular the topology used for CIFAR10 classification is rather large for the given task. We decided to use the same topologies as the state-of-the-art to allow for better comparison. In an ANN implementation, it is generally undesirable to use a network with a large number of parameters, since it increases the need for memory and computation. The relatively large number of parameters may to a certain extent explain the very low number of relative synaptic operations we observed during backpropagation. In an SNN, a large number of parameters is however less problematic from a computational perspective, since only the neurons which are activated by input spikes will trigger computations. A large portion of the network will therefore remain inactive. It would still be interesting to investigate signal sparsity and performance of \textit{SpikeGrad} in ANN topologies that were explicitly designed for minimal computation and memory requirements.

\newpage

\bibliographystyle{abbrv}

\end{document}